\documentclass[10pt,twocolumn,letterpaper]{article}
\usepackage[accsupp]{axessibility}  
\usepackage{iccv}
\usepackage{times}
\usepackage{epsfig}
\usepackage{graphicx}
\usepackage{amsmath}
\usepackage{amssymb}
\usepackage{booktabs}
\usepackage{multirow}
\usepackage{ulem}
\usepackage{color}
\usepackage{pifont}
\usepackage{balance}
\usepackage{colortbl}
\usepackage{xcolor}
\usepackage{makecell}
\usepackage{amsmath,amssymb,mathrsfs}
\usepackage{bm}
\usepackage{algorithm, algorithmic}
\newcommand{\cmark}{\ding{51}}

\usepackage{balance}
\usepackage{bbding} 
\usepackage{multirow}
\definecolor{newcolor}{rgb}{.8,.349,.1}

\usepackage[breaklinks=true,bookmarks=false]{hyperref}

\iccvfinalcopy 

\def\iccvPaperID{****} 

\ificcvfinal\fi

\begin{document}

\title{NDDepth: Normal-Distance Assisted Monocular Depth Estimation}

\author{Shuwei Shao$^{1,2}$\hspace{0.3in} 
	Zhongcai Pei$^{2,1}$\hspace{0.3in} 
	Weihai Chen$^{2,1}$\footnotemark[1]\hspace{0.3in}
	Xingming Wu$^{2,1}$\hspace{0.3in}
	Zhengguo Li$^{3}$\hspace{0.3in}\\
	$^1${School of Automation Science and Electrical Engineering, Beihang University, Beijing, China.}\\
	$^2${Hangzhou Innovation Institute, Beihang University, Hangzhou, Zhejiang, China}\\
	$^3$ {SRO department, Institute for Infocomm Research, A*STAR, Singapore} \\
	{\tt\small \{swshao, peizc, whchen\}@buaa.edu.cn}\hspace{0.3in}
	{\tt\small ezgli@i2r.a-star.edu.sg}\\
	\vspace{-2mm}
}

\maketitle
\ificcvfinal\thispagestyle{empty}\fi
\renewcommand{\thefootnote}{\fnsymbol{footnote}} 
\footnotetext[1]{Corresponding author} 
\begin{abstract}
	Monocular depth estimation has drawn widespread attention from the vision community due to its broad applications. In this paper, we propose a novel physics (geometry)-driven deep learning framework for monocular depth estimation by assuming that 3D scenes are constituted by piece-wise planes. Particularly, we introduce a new normal-distance head that outputs pixel-level surface normal and plane-to-origin distance for deriving depth at each position. Meanwhile, the normal and distance are regularized by a developed plane-aware consistency constraint. We further integrate an additional depth head to improve the robustness of the proposed framework. To fully exploit the strengths of these two heads, we develop an effective contrastive iterative refinement module that refines depth in a complementary manner according to the depth uncertainty. Extensive experiments indicate that the proposed method exceeds previous state-of-the-art competitors on the NYU-Depth-v2, KITTI and SUN RGB-D datasets. Notably, it \textbf{ranks 1st} among all submissions on the KITTI depth prediction online benchmark at the submission time. 
\end{abstract}

\normalem
\begin{figure}[!htb]
	\centering
	\includegraphics[width=0.9\linewidth]{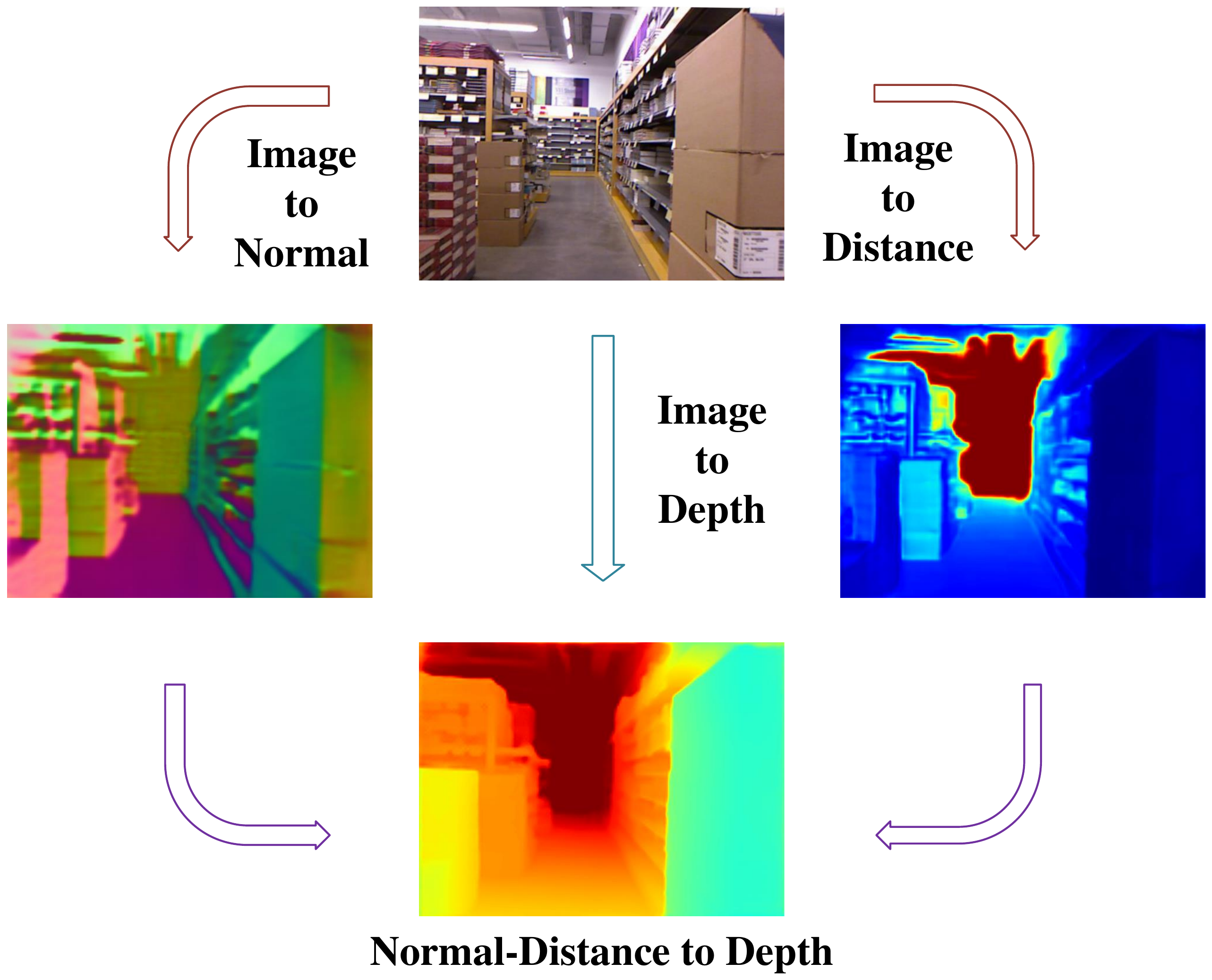}
	\caption{\textbf{A brief illustration of the relationship among image, surface normal, plane-to-origin distance and depth map}. }
	\label{Fig1}
\end{figure}
\section{Introduction}
Monocular depth estimation plays a more and more important role in computer vision, with applications ranging from robotics~\cite{tateno2017cnn,jia2022object}, scene understanding~\cite{hazirbas2016fusenet} to augmented reality~\cite{lee2011depth}. The task aims to estimate the depth map from a single RGB image. Considering that a 2D image can be projected from infinite number of 3D scenes, such a task is indeed ill-posed and inherently ambiguous. Hence, solving it reliably presents a formidable challenge for traditional methods~\cite{michels2005high,nagai2002hmm} due to their inherent limitations, typically involving low-dimensional and sparse distances or known and fixed objects.

Recently, much progress has been made in this field benefiting from the explosion of deep learning~\cite{eigen2014depth,fu2018deep,lee2019big,bhat2021adabins,Yuan_2022_CVPR}. Most efforts focus on designing increasingly complicated and powerful networks, which renders the task a hard fitting problem without the help of additional guidance. We argue that real-world 3D scenes usually have a high degree of regularity and proper scene priors should be incorporated into the framework to improve the nature of the solution. Planes are a common representation for modeling geometric prior knowledge of 3D scenes~\cite{bodis2014fast,chauve2010robust,liu2019planercnn}. A recent work by Patil~\textit{et al.}~\cite{patil2022p3depth} introduced a piece-wise planarity prior and used the offset vector field to borrow information from co-planar pixels. However, there is no direct constraint imposed on the offset vector field to help it learn about planar regions in their method. In addition, the planarity prior tends to fail in high-curvature regions, for example, bushes, trees, and other clutter, inevitably deteriorating the depth accuracy.

\begin{figure*}[!htb]
	\centering
	\includegraphics[width=0.9\linewidth]{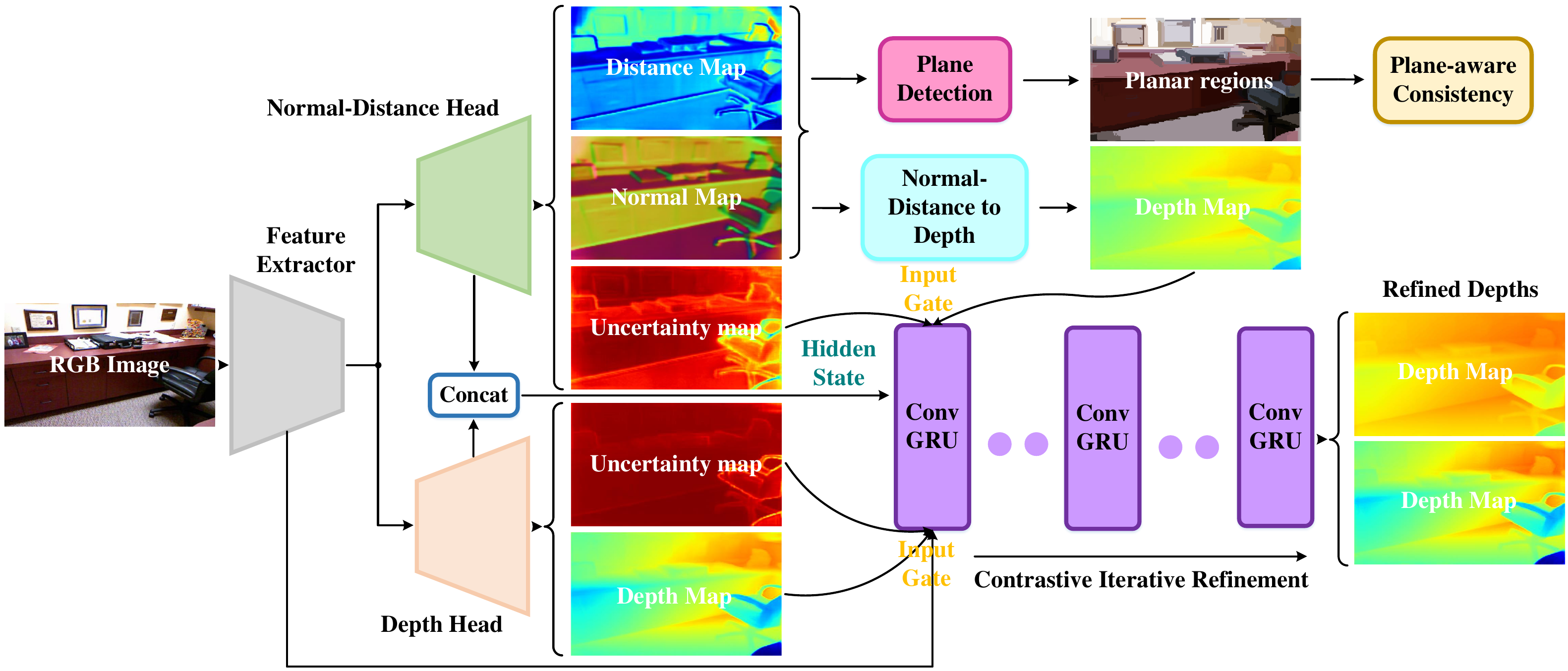}
	\caption{\textbf{Overview of the whole framework}. Note that the contrastive iterative refinement is performed at 1/4 resolution and the refined depth maps are upsampled to the full resolution using bilinear interpolation. For simplicity, we omit this detail in the flowchart.  }
	\label{Fig2}
\end{figure*}

In this paper, we propose a novel physics or geometry-driven deep learning framework to estimate a high-quality depth map from a single RGB image, by assuming that 3D scenes are  constituted with piece-wise planes. To be more specific, we parametrize the plane representation via surface normal and plane-to-origin distance (the distance from the associated plane to the origin, i.e., camera center in our case), which are predicted by our introduced normal-distance head. Furthermore, a plane-aware consistency constraint is developed to encourage the normal and distance to be piece-wise constant. To acquire planar regions, we adopt the Felzenszwalb segmentation algorithm~\cite{felzenszwalb2004efficient} for online plane detection using the geometric dissimilarity calculated from normal and distance. The detected planes could be noisy in early epochs because of inaccurate normal and distance predictions. Fortunately, they will gradually improve as the normal and distance quality improves, and in turn feedback to obtain better normal and distance predictions, which are afterwards converted to improved depth maps. It should be noted that the converted depth maps are prone to make severe errors in high-curvature regions due to the invalidity of the planarity assumption. To alleviate such failure cases, we integrate a second depth head designed in accordance with the regular paradigms,~\textit{e.g.},~\cite{Yuan_2022_CVPR}. In order to fully exploit the strengths of these two heads, uncertainty maps describing the depth uncertainty are additionally estimated. The depth and uncertainty maps are fed into a contrastive iterative refinement module for depth refinement in a complementary manner. The full pipeline is shown in Figure~\ref{Fig2}.

In the experiment, the proposed method is verified comprehensively on 3 standard datasets, including NYU-Depth-v2~\cite{silberman2012indoor}, KITTI~\cite{geiger2013vision} and SUN RGB-D~\cite{song2015sun}. Comparisons to leading approaches show that our method exceeds previous state-of-the-art competitors on the NYU-Depth-v2 and KITTI. In a challenging zero-shot setup, our method surpasses prior methods on the SUN RGB-D. Detailed ablation study is performed to demonstrate the merits of our developed components. In addition, we present qualitative comparisons with leading approaches to evidence the high quality of our depth and point cloud predictions.

To summarize, the contributions of this work are listed as follows:
\begin{itemize}
	\item We propose a new physics-driven deep learning framework for monocular depth estimation, which contains a normal-distance head and a depth head, built with asymmetric paradigms for depth acquisition.
	
	\item We introduce a plane-aware consistency constraint to regularize surface normal and plane-to-origin distance, and a contrastive iterative refinement module to refine depth in a complementary manner.
	
	\item The proposed method surpasses previous state-of-the-art approaches on the NYU-Depth-v2, KITTI and SUN RGB-D datasets, and \textbf{ranks 1st} on the KITTI depth prediction online benchmark at the submission time. 
\end{itemize}

\section{Related work}
\subsection{Monocular Depth Estimation}
Monocular depth estimation attempts to predict the depth map from a single RGB image. As one of the first learning-based studies, Saxena~\textit{et al.}~\cite{saxena2005learning} took into account both local and global features, and used a Markov Random Field to regress depth. Eigen~\textit{et al.}~\cite{eigen2014depth} pioneered the usage of convolutional neural network (CNN) in depth estimation, leveraging multi-scale networks for depth inference. Since then, many studies have sprung up to focus on this setting. Laina~\textit{et al.}~\cite{laina2016deeper} introduced a fully convolutional network based on residual learning and a reverse Huber loss for optimization. Cao~\textit{et al.}~\cite{cao2017estimating} and Fu~\textit{et al.}~\cite{fu2018deep} reformulated the depth regression problem as classification to predict easier depth ranges than the exact depth values. Lee~\textit{et al.}~\cite{lee2019big} designed several multi-scale guidance layers to establish the connection between intermediate layer features and the final depth map. Bhat~\textit{et al.}~\cite{bhat2021adabins} revisited the ordinal regression network~\cite{fu2018deep} and proposed to derive adaptive bins from the image content. Yang~\textit{et al.}~\cite{yang2021transformer} developed one of the first attempts of leveraging the Vision Transformer (ViT)~\cite{dosovitskiy2020image} to capture the long-distance correlation in depth estimation. In order to acquire the long-distance correlation in the decoder, Yuan~\textit{et al.}~\cite{Yuan_2022_CVPR} introduced neural window full-connected Conditional Random Fields (CRFs). Shao~\textit{et al.}~\cite{shao2023urcdc} adopted cross-distillation to exploit the strengths of Transformer and CNN. Nevertheless, most of these studies are data-driven approaches while the proposed physics-driven deep learning framework leverages the geometry of real-world 3D scenes. 

\subsection{Geometric Constraints for Depth}

Traditional methods for multi-view stereo~\cite{gallup2010piecewise} and 3D reconstruction~\cite{chauve2010robust,bodis2014fast} leveraged the planarity prior to enable faster optimization and address the challenges induced by poorly textured surfaces. Recently, Qi~\textit{et al.}~\cite{qi2018geonet} designed a geometric network to infer geometrically consistent surface normal and depth map. 
Long~\textit{et al.}~\cite{Long_2021_ICCV} developed an adaptive surface normal to regularize the depth map. 
Huynh~\textit{et al.}~\cite{huynh2020guiding} and Yin~\textit{et al.}~\cite{yin2019enforcing} introduced non-local coplanarity constraints either by the depth-attention module or the virtual normal. Kusupati~\textit{et al.}~\cite{kusupati2020normal} enforced a consistency
between spatial depth gradients calculated from estimated
depth and normal. Patil~\textit{et al.}~\cite{patil2022p3depth} proposed a piece-wise planarity prior and utilized the offset vector field to sample information from co-planar pixels. Unfortunately, the offset vector field is not given any direct constraints to aid in its comprehension of planar regions. Moreover, the planarity prior is not always valid, leading to erroneous depth estimates in high-curvature regions. By contrast, the planar constraint in our method is explicitly enforced inside each planar region, which avoids the interference of pixels from other planes. It is also worth noting that the depth head in our framework allows such failure cases in~\cite{patil2022p3depth} to be alleviated.

\section{Methodology}
Real-world 3D scenes usually have a high degree of regularity. It is therefore reasonable to assume that 3D scenes are constituted with piece-wise planes, and the surface normal and plane-to-origin distance are piece-wise constant. In this section, we introduce three main parts of the proposed framework in detail, including depth from normal-distance constraint, plane-aware consistency and contrastive iterative refinement.

\subsection{Depth from Normal-Distance Constraint}
Let ${\bf{P}} = {\left[ {X,Y,Z} \right]^{\rm{T}}}$ be a 3D point and $\textbf{p} = {\left[ {u,v} \right]^{\rm{T}}}$ be its projected 2D point on the image plane, corresponding to a planar region of the 3D scenes. The surface normal $\textbf{N}\left( \textbf{p} \right)$ is defined as the vector starting from $\textbf{P}$ and perpendicular to the associated plane. The plane-to-origin distance $\mathcal{D}\left( \textbf{p} \right)$ is defined as the distance between the associated plane and the origin (camera center in our case). The normal-distance constraint is formulated  as
\begin{equation} \textbf{N}\left( \textbf{p} \right)\textbf{P} = \mathcal{D}\left( \textbf{p} \right). \label{eq1} \end{equation}

According to the imaging principles of a pinhole camera, the projection from the 3D point $\textbf{P}$ to the 2D point $\textbf{p}$ is mathematically described as 
\begin{equation} \textbf{D}\left( \textbf{p} \right)\widetilde {\textbf{p}} = {\textbf{K}}\textbf{P}, \label{eq2} \end{equation} 
where $\textbf{D}\left( \textbf{p} \right)$ is the depth at $\textbf{p}$, $ \widetilde {\textbf{p}} $ denotes the homogeneous coordinate of $\textbf{p}$, ${\textbf{K}}$ denotes the intrinsic matrix.

By substituting Eq.~\ref{eq2} into Eq.~\ref{eq1}, we can acquire the depth via 
\begin{equation} \textbf{D}\left( \textbf{p} \right) = \frac{\mathcal{D}\left( \textbf{p} \right)}{{\textbf{N}\left( \textbf{p} \right){\textbf{K}^{ - 1}}\widetilde {\textbf{p}}}}. \label{eq4} \end{equation}

We design a normal-distance head to predict normal and distance, and then applies Eq.~\ref{eq4} to acquire a depth prediction. Compared with the direct prediction of depth, the intermediate normal-distance representation enjoys the benefit of piece-wise constant (not suitable for depth), which enables the plane-aware consistency constraint to establish interactions among pixels and thus results in improved accuracy of depth estimate.

\subsection{Plane-aware Consistency}
\textbf{Planar region detection}. In order to enforce the plane-aware consistency constraint, we need to detect the planar regions correctly. Following previous works~\cite{concha2014using,concha2015dpptam,yu2020p,li2021structdepth}, we adopt the Felzenszwalb segmentation algorithm~\cite{felzenszwalb2004efficient} in our approach. 

\begin{figure}[!htb]
	\centering
	\includegraphics[width=1.0\linewidth]{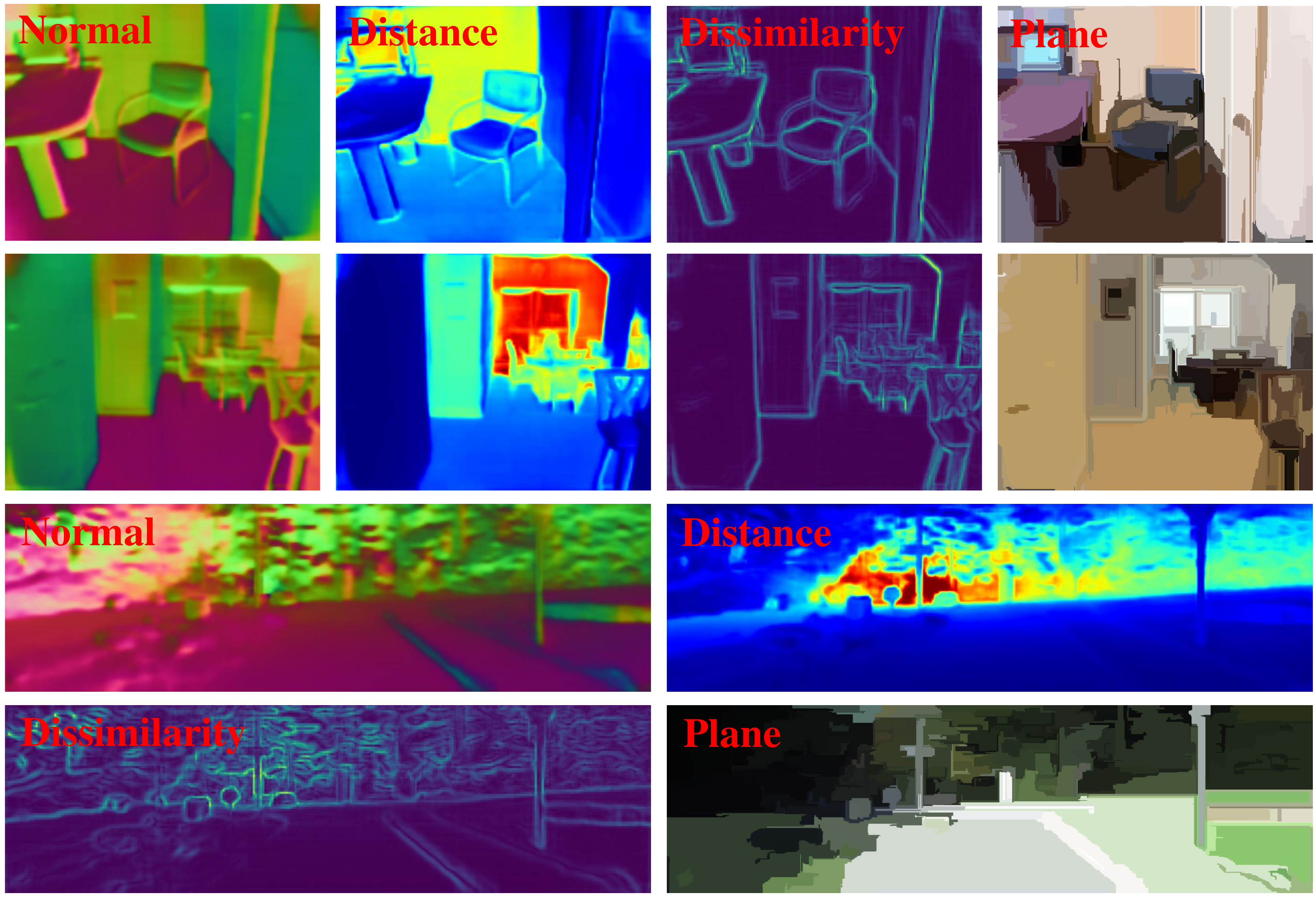}
	\caption{\textbf{Visualization of surface normal, plane-to-origin distance, geometric dissimilarity and extracted planar regions.}  }
	\label{Fig4}
\end{figure}

Let \textbf{q} be an adjacent pixel of \textbf{p}. We define the normal dissimilarity between them as
\begin{equation}di{s_\textbf{N}}\left( {\textbf{p},\textbf{q}} \right) = \left\| {\textbf{N}(\textbf{q}) - \textbf{N}(\textbf{p})}\right\|,\end{equation}
where $\left\|  \cdot  \right\|$ denotes the Euclidean distance. Suppose $dis_\textbf{N}^{\max }$ and $dis_\textbf{N}^{\min }$ are the maximum and minimum dissimilarities among all adjacent pixels, which we use to normalize the dissimilarity via
\begin{equation}{\overline {dis} _\textbf{N}} = \frac{{di{s_\textbf{N}}\left( {\textbf{p},\textbf{q}} \right) - dis_\textbf{N}^{\min }}}{{dis_\textbf{N}^{\max } - dis_\textbf{N}^{\min }}}.\end{equation}
The distance dissimilarity is computed as 
\begin{equation}di{s_\mathcal{D}}\left( {\textbf{p},\textbf{q}} \right) = \left| {{\mathcal{D}(\textbf{q}) - \mathcal{D}(\textbf{p})}} \right|. \end{equation}
The geometric dissimilarity is a combination of normalized normal and distance dissimilarities
\begin{equation}{\overline {dis} _g} = {\overline {dis} _\textbf{N}} + {\overline {dis} _\mathcal{D}}.\end{equation}

Based on the geometric dissimilarity, we use the Felzenszwalb segmentation algorithm to perform online plane detection. Intuitively, the planar regions, for example, floors and walls in indoor scenarios and roads in outdoor scenarios, are more likely to lie within a larger area, and we only select regions larger than 200 pixels. In Figure~\ref{Fig4}, we show some qualitative results of normal, distance, geometric dissimilarity and detected planes. 

\textbf{Plane-aware consistency loss}. After detection of planar regions, we encourage plane-aware consistency by penalizing the first-order gradients of normal and distance within planar regions $\mathcal{M}\left( \textbf{p} \right)$
\begin{equation} \resizebox{0.89\hsize}{!}{${\mathcal{L}_{pc}} = \sum\limits_\textbf{p} {\mathcal{M}\left( \textbf{p} \right){{ \left|\nabla\textbf{N}\left( \textbf{p} \right) \right|}}} + \sum\limits_\textbf{p} {\mathcal{M}\left( \textbf{p} \right) {{ \left|\nabla\mathcal{D}\left( \textbf{p} \right) \right|}} }$} . \end{equation}
While the initial normal and distance predictions may not be precise, they will gradually improve as the training epochs, leading to a better segmentation and vice versa. 

The depth maps converted from normals and distances are not always working, and large errors are prone to occur in high-curvature regions. We make an observation that depth maps derived from regular paradigms have smaller errors in these regions. Hence, the work in ~\cite{Yuan_2022_CVPR} is leveraged to define a second head of our framework. To fully exploit the strengths of these two heads, we model the depth uncertainty and develop a contrastive iterative refinement module to refine depth maps in a complementary manner.
\subsection{Contrastive Iterative Refinement}
\textbf{Uncertainty modeling.} The areas with large errors need to be identified before performing the depth refinement. We model the depth uncertainty with a function inspired by the probability density function of Laplace distribution~\cite{kendall2017uncertainties}
\begin{equation}
	\textbf{U}^{gt}\left( \textbf{p} \right) = 1 - \exp \left( { - \frac{{\left| {{\textbf{D}}\left( \textbf{p} \right) - \textbf{D}^{gt}\left( \textbf{p} \right)} \right|}}{b }} \right),
\end{equation}
where $\textbf{D}^{gt}\left( \textbf{p} \right)$ denotes the ground-truth depth map, $b$ is a coefficient that controls the error tolerance. Since $\textbf{D}^{gt}\left( \textbf{p} \right)$ is not available at test time, uncertainty maps are additionally estimated to approximate ${\textbf{U}_1^{gt}} \left( \textbf{p} \right) $ and ${\textbf{U}_2^{gt}} \left( \textbf{p} \right)$, which is supervised by 
\begin{equation}\resizebox{0.87\hsize}{!}{${\mathcal{L}_{\textbf{U}}} = \sum\limits_\textbf{p} {\left| {\textbf{U}_1\left( \textbf{p} \right) - {\textbf{U}_1^{gt}} \left( \textbf{p} \right)} \right|}  + \sum\limits_\textbf{p} {\left| {\textbf{U}_2\left( \textbf{p} \right) - {\textbf{U}_2^{gt}} \left( \textbf{p} \right)} \right|}$}, 
\end{equation}
where $\textbf{U}_1\left( \textbf{p} \right)$ and $\textbf{U}_2\left( \textbf{p} \right)$ denote the uncertainty predictions from normal-distance and depth heads, respectively.

Aside from the uncertainty map, we calculate the absolute difference between $\textbf{D}_1 \left( \textbf{p} \right)$ and $\textbf{D}_2 \left( \textbf{p} \right)$ to indicate the complementary regions in these two depth maps, i.e., complementary map,
\begin{equation}di{f_\textbf{D}}\left( \textbf{p} \right) = \left| {\textbf{D}_1 \left( \textbf{p} \right) - \textbf{D}_2 \left( \textbf{p} \right)} \right|, \end{equation}
where $\textbf{D}_1 \left( \textbf{p} \right)$ and $\textbf{D}_2 \left( \textbf{p} \right)$ denote the depth predictions from normal-distance and depth heads, respectively. 

\textbf{Contrastive iterative update.}
Grounded on the uncertainty map and complementary map, we refine depth maps iteratively. At each iteration, we update $\textbf{D}_1 \left( \textbf{p} \right)$ and $\textbf{D}_2 \left( \textbf{p} \right)$ as
\begin{equation} \textbf{D}_1^{t + 1} \leftarrow \textbf{D}_1^t + \Delta \textbf{D}_1^t,\textbf{D}_2^{t + 1} \leftarrow \textbf{D}_2^t + \Delta \textbf{D}_2^t, \end{equation}
where $\Delta \textbf{D}_1^t$ and $\Delta \textbf{D}_2^t$ denote the updates at iteration $t$. Inspired by~\cite{teed2020raft}, we adopt a convolutional gated recurrent unit (ConvGRU)~\cite{chung2014empirical} to yield these updates, since the ConvGRU can memorize the history status and make full use of the temporal information during the refinement process. The ConvGRU operates at 1/4 resolution.

\begin{table*}[htb!]
	\begin{center}
		\renewcommand{\arraystretch}{1.3}
		\resizebox{1.68\columnwidth}{!}{\begin{tabular}{c || c || c c c c|| c c c}	
				\Xhline{1.2pt}
				Method &  Cap & Abs Rel $\downarrow$ & Sq Rel $\downarrow$ & RMSE $\downarrow$ & ${\textbf{\rm{log}}_{\bm{{10}}}}$ $\downarrow$ &  $\delta  < 1.25$ $\uparrow$ &  $\delta  < {1.25^2}$ $\uparrow$& $\delta  < {1.25^3}$ $\uparrow$ \\
				\hline						
				\hline
				Fu~\textit{et al.}~\cite{fu2018deep}& 0-10m&0.115&-&0.509&0.051&0.828&0.965&0.992
				\\
				VNL~\cite{yin2019enforcing}& 0-10m&0.108&-&0.416&0.048&0.875&0.976&0.994
				\\
				BTS~\cite{lee2019big}& 0-10m&0.113&0.066&0.407&0.049&0.871&0.977&0.995
				\\
				DAV~\cite{huynh2020guiding}& 0-10m&0.108&-&0.412&-&0.882&0.980&0.996
				\\
				PWA~\cite{lee2021patch}& 0-10m&0.105&-&0.374&0.045&0.892&0.985&0.997
				\\
				Long~\textit{et al.}~\cite{Long_2021_ICCV}& 0-10m&0.101&-&0.377&0.044&0.890&0.982&0.996
				\\
				TransDepth~\cite{yang2021transformer}& 0-10m&0.106&-&0.365&0.045&0.900&0.983&0.996
				\\
				Adabins~\cite{bhat2021adabins}& 0-10m&0.103&-&0.364&0.044&0.903&0.984&0.997
				\\
				P3Depth~\cite{patil2022p3depth}& 0-10m&0.104&-&0.356&0.043&0.898&0.981&0.996
				\\ 	
				NeWCRFs~\cite{Yuan_2022_CVPR}& 0-10m&0.095&0.045&0.334&0.041&0.922&\textbf{0.992}&\textbf{0.998}
				\\ 					
				\hline
				\textbf{Ours} & 0-10m&\textbf{0.087} &\textbf{0.041}&\textbf{0.311}&\textbf{0.038}&\textbf{0.936}&0.991&\textbf{0.998}\\
				\Xhline{1.2pt}
		\end{tabular}}
	\end{center}
	\caption{\textbf{Quantitative depth comparison on the NYU-Depth-v2 dataset}. } 
\label{table1}
\end{table*}

Specifically, we first project $\textbf{D}_1^t$, $\textbf{D}_2^t$, $\textbf{U}_1$, $\textbf{U}_2$ and $dif_\textbf{D}^t$ into the feature space using two convolutional layers. Then, we concatenate the projected feature and the image contextual feature (1/4 resolution output in the feature extractor) to form a tensor $\textbf{I}^t$ as the input. The structure inside ConvGRU is as follows
\begin{equation}
	{{\bm{{\rm z}}}^{t+1}} = \sigma \left( {Con{v_{5 \times 5}}\left( {\left[ {{{\bm{{\rm h}}}^{t}}, \textbf{I}^t} \right], W_z} \right)} \right)
\end{equation}
\begin{equation}
	{{\bm{{\rm r}}}^{t+1}} = \sigma \left( {Con{v_{5 \times 5}}\left( {\left[ {{{\bm{{\rm h}}}^{t}}, \textbf{I}^t} \right]}, W_r\right)} \right)
\end{equation}
\begin{equation}
	{\widehat {\bm{{\rm h}}}^{t+1}} = \tanh \left( {Con{v_{5 \times 5}}\left( {\left[ {{{\bm{{\rm r}}}^{t+1}} \odot {{\bm{{\rm h}}}^{t}}, \textbf{I}^t}\right]}, W_h  \right)} \right)
\end{equation}
\begin{equation}
	{{\bm{{\rm h}}}^{t+1}} = \left( {1 - {{\bm{{\rm z}}}^{t+1}}} \right) \odot {{\bm{{\rm h}}}^{t}} + {{\bm{{\rm z}}}^{t+1}} \odot {\widehat {\bm{{\rm h}}}^{t+1}},
\end{equation}
where $Con{v_{5 \times 5}}$ stands for the separable ${5 \times 5}$ convolution, $\odot$ denotes the element-wise multiplication, $\sigma$ and tanh denote the sigmoid and tanh activation functions. The hidden state ${{\bm{{\rm h}}}^{t}}$ is initialized by the concatenated penultimate layer feature maps of the normal-distance and depth heads, with the tanh function as activation.

With this refinement module, starting from an initial solution, the depth maps are iteratively refined and eventually converge to the final results as $\textbf{D}_1^{*} \leftarrow \textbf{D}_1^t,\textbf{D}_2^{*} \leftarrow \textbf{D}_2^t$. Finally, we upsample $\textbf{D}_1^{*}$ and $\textbf{D}_2^{*}$ to the full resolution, and average them as the output of the whole framework,
\begin{equation}
{\textbf{D}^*\left( \textbf{p} \right)} = 0.5(\textbf{D}_1^*\left( \textbf{p} \right) + \textbf{D}_2^*\left( \textbf{p} \right)).
\end{equation}

\subsection{Overall Loss and Architecture}

\textbf{Depth loss}. We adopt a
scaled Scale-Invariant loss for depth supervision~\cite{lee2019big},
\begin{equation}
\resizebox{0.85\hsize}{!}{${\mathcal{L}_\textbf{D}} = \sum\limits_{s = 1}^m {{\gamma ^{m - s}}} \left( \begin{array}{l}
		\kappa \sqrt {\frac{1}{{\left| {\bf{T}} \right|}}\sum\limits_{\bf{p}} {{{\left( {{{\bf{g}}_1}\left( {\bf{p}} \right)} \right)}^2} - \frac{\eta }{{{{\left| {\bf{T}} \right|}^2}}}{{\left( {\sum\limits_{\bf{p}} {{{\bf{g}}_1}\left( {\bf{p}} \right)} } \right)}^2}} }  + \\
		\kappa \sqrt {\frac{1}{{\left| {\bf{T}} \right|}}\sum\limits_{\bf{p}} {{{\left( {{{\bf{g}}_2}\left( {\bf{p}} \right)} \right)}^2} - \frac{\eta }{{{{\left| {\bf{T}} \right|}^2}}}{{\left( {\sum\limits_{\bf{p}} {{{\bf{g}}_2}\left( {\bf{p}} \right)} } \right)}^2}} } 
	\end{array} \right)$},
\end{equation}
where $\gamma$ is the decay factor and $m$ is the maximum iteration step, set as 0.85 and 3, respectively, ${\textbf{g}}\left( \textbf{p} \right) = \log {\textbf{D}}\left( \textbf{p} \right) - \log \textbf{D}^{gt}\left( \textbf{p} \right)$, \textbf{T} stands for a collection of pixels with valid values, $\kappa$ and $\eta$ are selected as 10 and 0.85 based on~\cite{lee2019big}.

\textbf{Normal loss}. We adopt a negative cosine loss for normal supervision~\cite{eigen2015predicting},
\begin{equation} {\mathcal{L}_\textbf{N}} = \sum\limits_\textbf{p} {1 - \textbf{N}\left( \textbf{p} \right) {\textbf{N}^{gt}}^{\rm{T}}\left( \textbf{p} \right)},  
\end{equation}

\textbf{Distance loss}. We adopt an L1 loss for distance supervision,
\begin{equation}
{\mathcal{L}_\mathcal{D}} = \sum\limits_\textbf{p} {\left| {\mathcal{D}\left( \textbf{p} \right) - {\mathcal{D}^{gt}}\left( \textbf{p} \right)} \right|}, \end{equation}

As the NYU-Depth-v2 and KITTI datasets do not contain normal and distance ground-truth, we acquire the normal ground-truth following~\cite{qiu2019deeplidar}. Then, we acquire the distance ground-truth by  ${\cal D}^{gt}\left( {\bf{p}} \right) = {\bf{D}}\left( {\bf{p}} \right)^{gt} {\bf{N}}\left( {\bf{p}} \right)^{gt}{{\bf{K}}^{ - 1}}{\widetilde {\textbf{p}}}$.

\textbf{Overall loss.} We define the overall loss as
\begin{equation}{\mathcal{L}_{overall = }}{\lambda _1}{\mathcal{L}_\textbf{D}} + {\lambda _2}{\mathcal{L}_\textbf{N}} + {\lambda _3}{\mathcal{L}_\mathcal{D}} + {\lambda _4}{\mathcal{L}_\textbf{U}} + {\lambda _5}{\mathcal{L}_{pc}},
\end{equation}
where ${\lambda _1}$, ${\lambda _2}$, ${\lambda _3}$, ${\lambda _4}$ and ${\lambda _5}$ are empirically set as 1, 5, 0.25, 1 and 0.01, respectively. 

\textbf{Network architecture.} The feature extractor adopts the Swin-L~\cite{Liu_2021_ICCV} when not otherwise specified. For KITTI official split with more data, the feature extractor is SwinV2-L~\cite{liu2022swin} that has a larger window size. The depth head follows the decoder design of NeWCRFs~\cite{Yuan_2022_CVPR}. The normal-distance head shares a same architecture as depth head apart from the final layer, which is devised to estimate normal and distance jointly.

\begin{table*}[htb!]
\begin{center}
	\renewcommand{\arraystretch}{1.3}
	\resizebox{1.70\columnwidth}{!}{\begin{tabular}{c|| c || c c  c c || c c c }	
			\Xhline{1.2pt}
			Method & Cap & Abs Rel $\downarrow$ & Sq Rel $\downarrow$ & RMSE $\downarrow$ & RMSE log $\downarrow$ & $\delta  < 1.25$ $\uparrow$ & $\delta  < {1.25^2}$ $\uparrow$& $\delta  < {1.25^3}$ $\uparrow$ \\
			\hline						
			\hline
			VNL~\cite{yin2019enforcing}&0-80m&0.072&-&3.258&0.117&0.938&0.990&0.998
			\\
			BTS~\cite{lee2019big}&0-80m&0.061&0.261&2.834&0.099&0.954&0.992&0.998
			\\
			PWA~\cite{lee2021patch}&0-80m&0.060&0.221&2.604&0.093&0.958&0.994&\textbf{0.999}
			\\
			TransDepth~\cite{yang2021transformer}&0-80m&0.064&0.252&2.755&0.098&0.956&0.994&\textbf{0.999}
			\\
			Adabins~\cite{bhat2021adabins}&0-80m&0.058&0.190&2.360&0.088&0.964&0.995&\textbf{0.999}
			\\ 		
			P3Depth~\cite{patil2022p3depth}&0-80m&0.071&0.270&2.842&0.103&0.953&0.993&0.998
			\\
			NeWCRFs~\cite{Yuan_2022_CVPR}&0-80m&0.052&0.155&2.129&0.079&0.974&0.997&\textbf{0.999}
			\\				
			\hline
			\textbf{Ours} &0-80m &\textbf{0.050}&\textbf{0.141}&\textbf{2.025}&\textbf{0.075}&\textbf{0.978}&\textbf{0.998}&\textbf{0.999}\\
			\hline
			\hline
			BTS~\cite{lee2019big}&0-50m&0.058&0.183&1.995&0.090&0.962&0.994&0.999
			\\
			PWA~\cite{lee2021patch}&0-50m&0.057&0.161&1.872&0.087&0.965&0.995&0.999
			\\
			P3Depth~\cite{patil2022p3depth}&0-50m&0.055&0.130&1.651&0.081&0.974&0.997&0.999
			\\
			\hline
			\textbf{Ours} &0-50m&\textbf{0.048}&\textbf{0.107}&\textbf{1.513}&\textbf{0.071}&\textbf{0.981}&\textbf{0.998} &\textbf{1.000}\\
			\Xhline{1.2pt}
			
	\end{tabular}}
\end{center}
\caption{\textbf{Quantitative depth comparison on the Eigen split of KITTI dataset}. ``-'' indicates not applicable. The best results are marked in \textbf{bold}.}
\label{table2}
\end{table*}

\begin{table}[htb!]
	\begin{center}
		\renewcommand{\arraystretch}{1.3}
		\resizebox{0.93\columnwidth}{!}{\begin{tabular}{c|| c c c c  }	
				\Xhline{1.2pt}
				Method & SILog $\downarrow$ & Sq Rel $\downarrow$ & Abs Rel $\downarrow$ & iRMSE$\downarrow$\\
				\hline						
				\hline
				P3Depth~\cite{patil2022p3depth}&12.82&9.92&2.53&13.71
				\\
				VNL~\cite{yin2019enforcing}&12.65&10.15&2.46&13.02
				\\
				Fu~\textit{et al.}~\cite{fu2018deep}&11.77&8.78&2.23&12.98
				\\
				BTS~\cite{lee2019big}&11.67&9.04&2.21&12.23
				\\
				BA-Full~\cite{aich2021bidirectional}&11.61&9.38&2.29&12.23
				\\
				PackNet-SAN~\cite{guizilini2021sparse}&11.54&9.12&2.35&12.38
				\\
				PWA~\cite{lee2021patch}&11.45&9.05&2.30&12.32
				\\
				NeWCRFs~\cite{Yuan_2022_CVPR}&10.39&8.37&1.83&11.03
				\\
				\hline
				\textbf{Ours} &\textbf{9.62} &\textbf{7.75}&\textbf{1.59}&\textbf{10.62}
				\\
				\Xhline{1.2pt}		
		\end{tabular}}
	\end{center}
	\caption{\textbf{Quantitative depth comparison on the official split of KITTI dataset}. Note that the Sq Rel is calculated in a different way here. The SILog is the main ranking metric.} 
\label{table3}
\end{table}

\section{Experiment}
We conduct comprehensive experiments on datasets for indoor and outdoor scenes, which includes the NYU-Depth-v2, KITTI and SUN RGB-D. In this section, we first give a detailed description of the relevant datasets, evaluation metrics and implementation details. Then, we provide quantitative and qualitative comparisons to previous state-of-the-art competitors. Finally, we demonstrate zero-shot generalization and ablation studies to validate its generalizability and the efficacy of each key component.

\subsection{Datasets and Evaluation Metrics}
\textbf{NYU-Depth-v2 dataset} is collected from indoor scenarios and provides images and ground-truth depth maps at a resolution of $640 \times 480$ pixels. We adopt the official data split to evaluate our method. The split employs 249 scenes for training and 654 images from 215 scenes for testing.

\textbf{KITTI dataset} is acquired from outdoor scenes using camera and LiDAR mounted on a moving vehicle, and consists of stereo images and 3D laser scans. The image resolution is around $1241 \times 376$ pixels. Here we adopt two mainly used data splits. One is the Eigen split with 23488 training images and 697 test images~\cite{eigen2014depth}. The other one is the official split with 85898 training images, 1000 validation images and 500 test images. For the official split, the ground-truth depth maps of the test images are not available, and the results are provided by the online server.

\textbf{SUN RGB-D dataset} is captured from multiple indoor scenes with four sensors, and contains roughly 10K images. We only use this dataset for generalization study in a challenging zero-shot setup. The pre-trained models are evaluated on the official test set of 5050 images.

\textbf{Evaluation Metrics.} Following~\cite{Yuan_2022_CVPR}, we adopt the standard evaluation metrics in our experiments.

\begin{figure*}[!htb]
\centering
\includegraphics[width=0.97\linewidth]{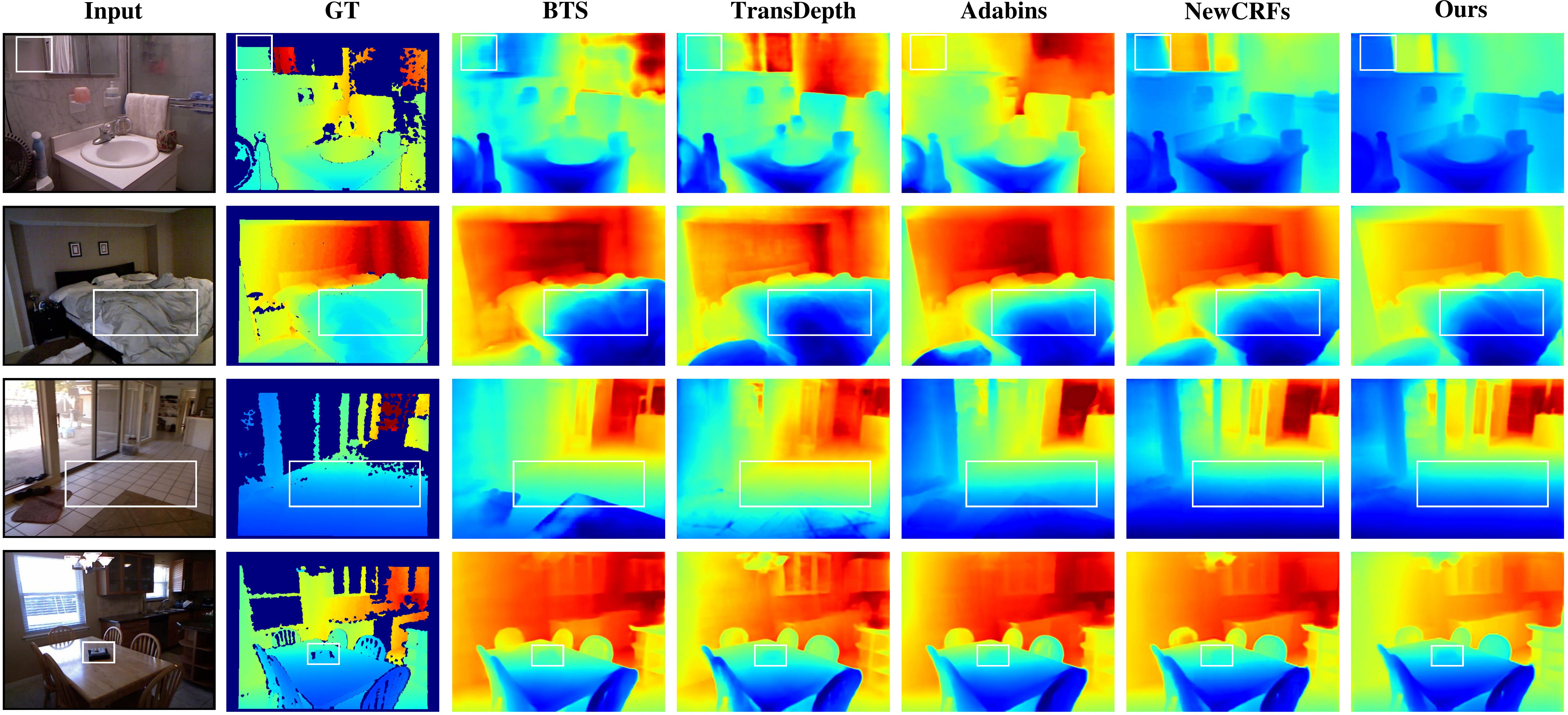}
\caption{\textbf{Qualitative depth results on the NYU-Depth-v2 dataset}. The white boxes highlight the regions to emphasize. }
\label{Fig6}
\end{figure*}

\begin{figure*}[!htb]
\centering
\includegraphics[width=0.97\linewidth]{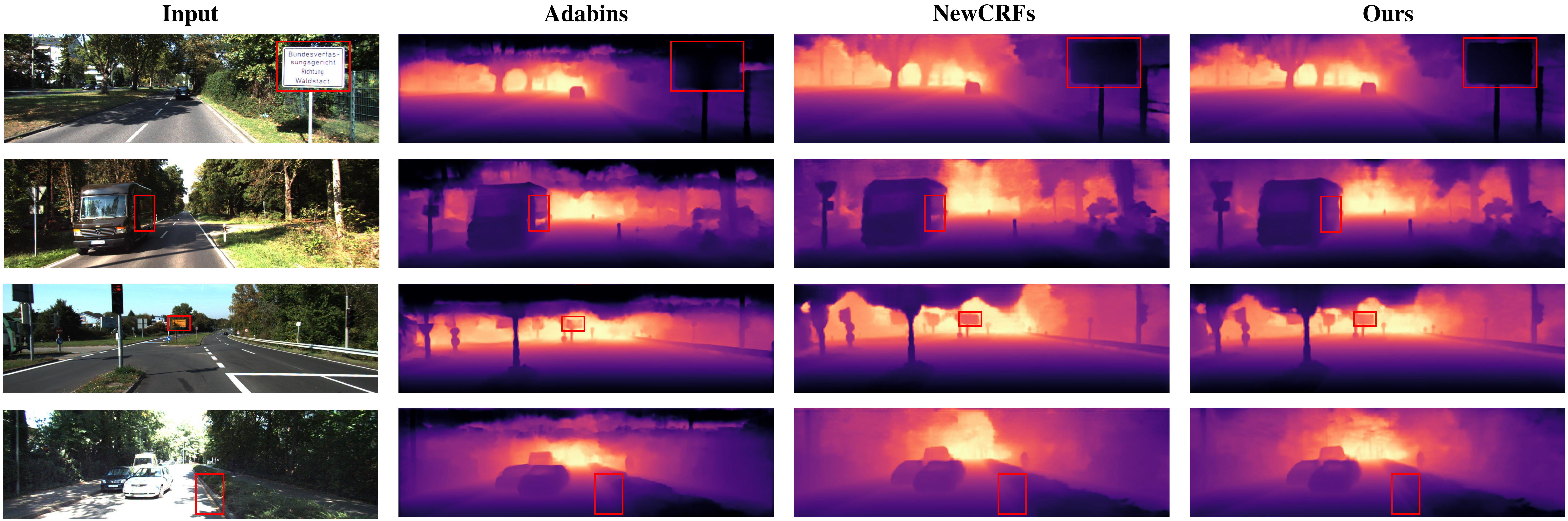}
\caption{\textbf{Qualitative depth results on the KITTI dataset}. The red boxes highlight the regions to emphasize. }
\label{Fig5}
\end{figure*}

\begin{figure}[!htb]
\centering
\includegraphics[width=1.0\linewidth]{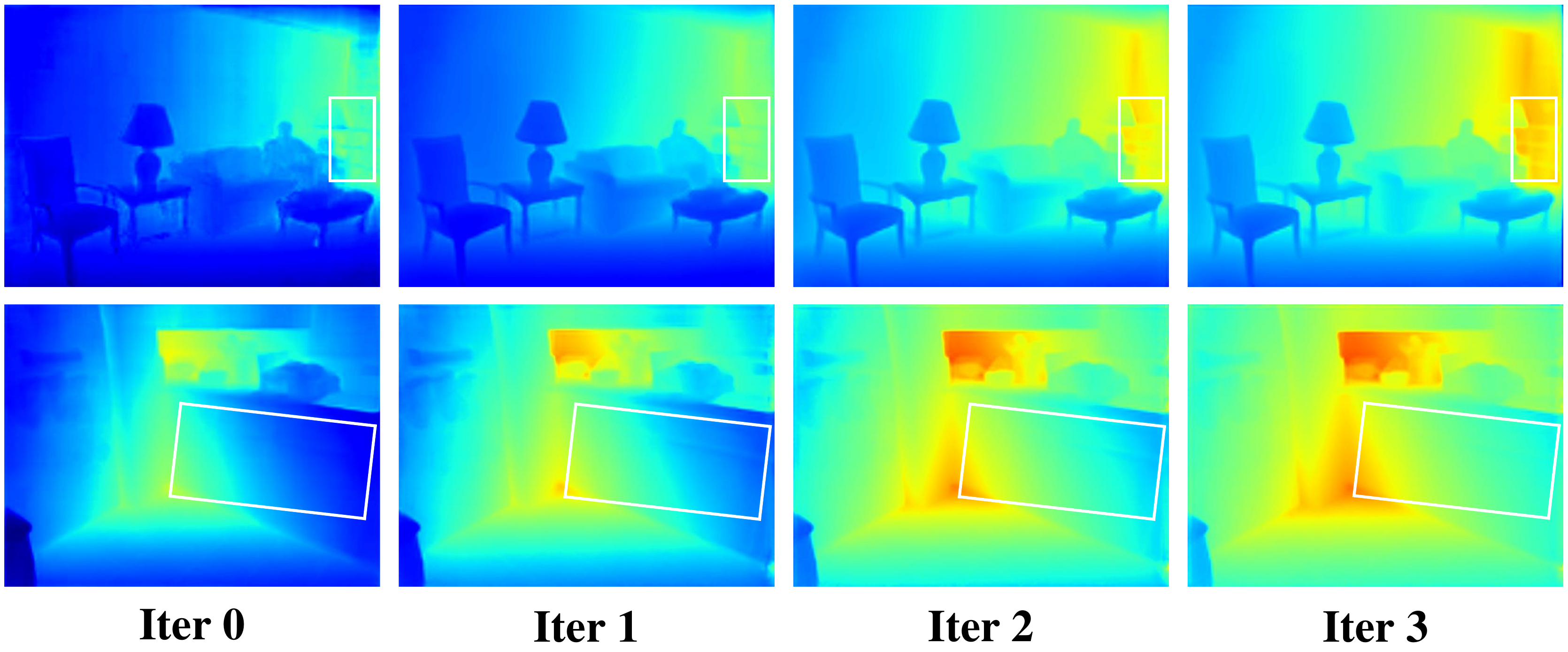}
\caption{\textbf{Visualization of the refinement process in the contrastive iterative refinement module}. The first depth map of the upper row at iter 0 is from the normal-distance head and the first depth map of the lower row at iter 0 is from the depth head. The white boxes show the regions to focus on.}
\label{Fig7}
\end{figure}

\begin{figure}[!htb]
\centering
\includegraphics[width=1.0\linewidth]{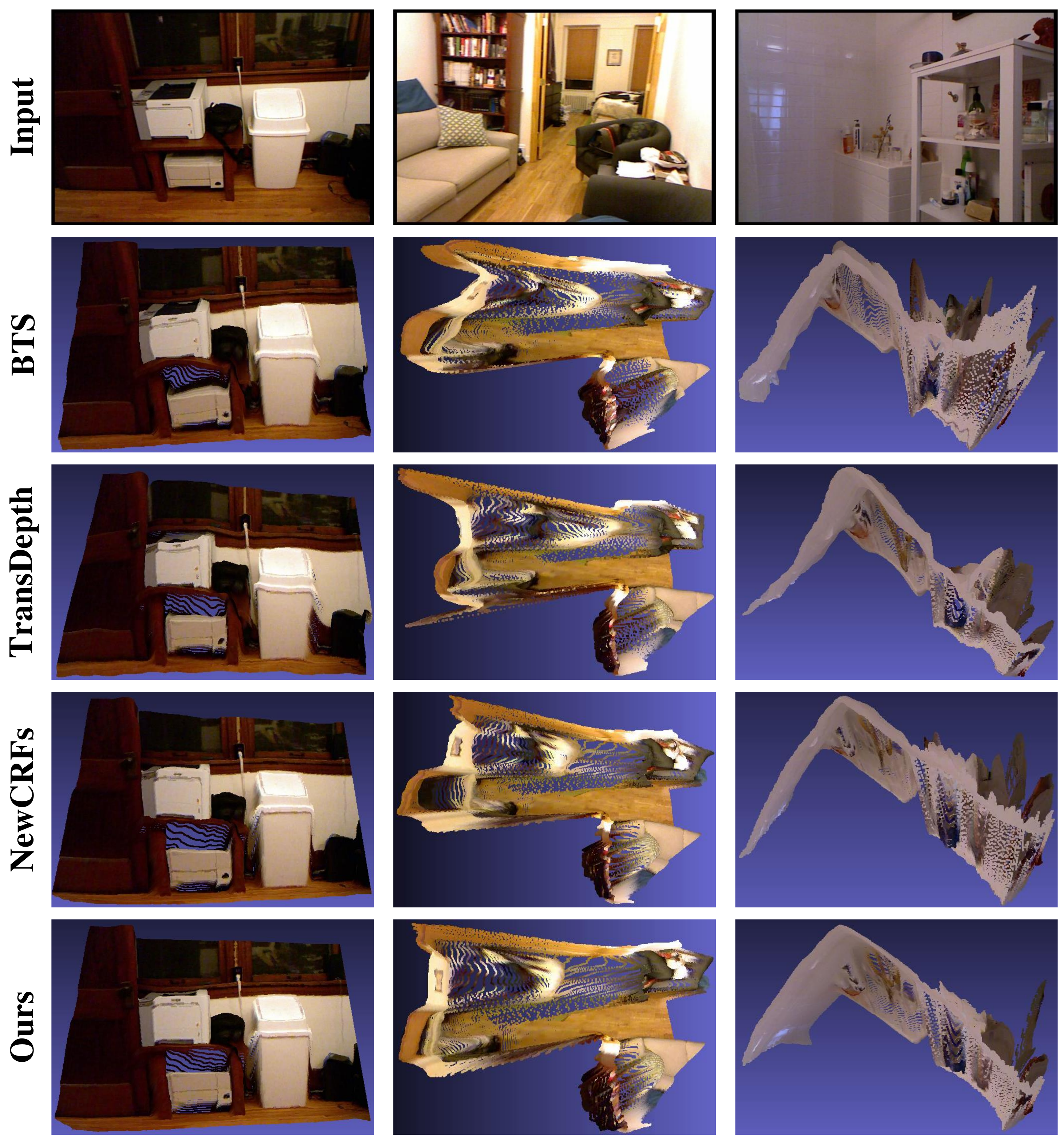}
\caption{\textbf{Qualitative point cloud results on the NYU-Depth-v2 dataset.} The point clouds in the first column are captured from the front view, and the point clouds in the second and third columns are captured from the top view. }
\label{Fig8}
\end{figure}
\subsection{Implementation Details}
Our framework is implemented in PyTorch~\cite{paszke2017automatic} and experimented on NVIDIA RTX A5000 GPUs. We use the Adam optimizer~\cite{kingma2015adam} where $\beta_1=0.9, \beta_2=0.999$ and a batch size of 8. The learning rate is scheduled through polynomial decay from a base value of 2e-5 to 2e-6. The training process runs a total number of 25 epochs.

\subsection{Comparison to Prior Competitors}
\textbf{NYU-Depth-v2.} As summarized in Table~\ref{table1}, our method
surpasses previous competing approaches on most metrics,
especially notable are results on Abs Rel and RMSE, which
emphasizes our contributions in boosting the performance.
In Figure~\ref{Fig6}, we demonstrate qualitative depth comparisons. As
can be seen, our method is better at delineating planar regions while preserving local details, such as edges, and has
high immunity against texture variations. Furthermore, we
convert depth maps into point clouds, and show qualitative
point cloud comparisons in Figure~\ref{Fig8}. Our point clouds have
fewer distortions than others. The point clouds of other approaches suffer from severe distortions and struggle to preserve prominent geometric features. 

\textbf{KITTI.} To further show the applicability of our method
in the outdoor scenario, we evaluate it on the KITTI
dataset. We first perform comparison on the Eigen split.
As listed in Table~\ref{table2}, our method exceeds previous leading approaches by a noticeable margin on most metrics, such as Sq Rel and RMSE, when the maximum depth is capped at 80m. Regarding the 0-50m cap, our method goes well beyond the P3Depth. Besides, we notice that compared to the PWA, the P3Depth performs better at the 0-50m cap but significantly worse at the 0-80m cap. This is due to the fact that most of the nearby parts are planar regions such as roads, while the
distant parts are occupied by high-curvature regions, such
as bushes, trees and other clutter, resulting in the planarity
prior to fail. Our method relaxes such failure cases with the
help of depth head and achieves promising results on both
caps. Figure~\ref{Fig5} presents qualitative depth comparisons. As can
be seen, our method is more capable of planar regions, e.g.,
sign board.

We then compare our method against prior leading approaches on the official split. The results are listed in Table~\ref{table3} and are available from the online server. Here we can
see that our method exceeds previous approaches again and
achieves consistent improvements on all metrics. Besides, the main ranking metric SILog is reduced markedly, and our method \textbf{ranks 1st} among all submissions on the KITTI depth prediction online benchmark at the submission time.

\subsection{Zero-shot Generalization}

In Table~\ref{table5}, we show generalization comparison in
a zero-shot setting where the test dataset has not been seen
during training. The models are trained on the NYU-Depthv2 dataset but evaluated on the SUN RGB-D dataset. The
superior results indicate that our physics-driven deep learning framework does learn transferable features rather than simply memorizing
statistics of training data.

\subsection{Ablation Study}
To better understand the effect of each key component in our framework, we conduct an ablation study and present the results in Table~\ref{table6}.

We find that the normal-distance head works better than the depth head in a standalone setting, which seems to be contrary to the finding in~\cite{patil2022p3depth} that directly predicting depth is better than predicting plane coefficients. The reason may be that compared with the implicit plane representation used in~\cite{patil2022p3depth}, we adopt a more explicit normal-distance representation, which allows us to enforce direct supervisory signals to guide the normal and distance learning, hence leading to more accurate depth estimates. By first enforcing the plane-aware consistency constraint to normal and distance, we observe a consistent improvement on most metrics. On the basis of the normal-distance head, we further integrate the depth head into the framework. The resulting improvement indicates that the depth estimates from these two heads are inherently complementary to each other. Finally, we add the contrastive iterative refinement module and form the complete framework, achieving the best results.

\begin{table}[htb!]
	\begin{center}
		\renewcommand{\arraystretch}{1.3}
		\resizebox{1.0\columnwidth}{!}{\begin{tabular}{c|| c c c c c  }	
				\Xhline{1.2pt}
				Method & Abs Rel $\downarrow$ & RMSE $\downarrow$ & ${\textbf{\rm{log}}_{\bm{{10}}}}$ $\downarrow$ &  $\delta  < 1.25$ $\uparrow$ &  $\delta  < {1.25^2}$ $\uparrow$ \\
				\hline						
				\hline
				Chen~\cite{chen2019structure}&0.166&0.494&0.071&0.757
				&0.943 \\
				VNL~\cite{yin2019enforcing}&0.183&0.541&0.082&0.696
				&0.912 \\
				BTS~\cite{lee2019big}&0.172&0.515&0.075&0.740
				&0.933 \\
				Adabins~\cite{bhat2021adabins}&0.159&0.476&0.068&0.771
				&0.944 \\
				\hline
				\textbf{Ours} &\textbf{0.137} &\textbf{0.411}&\textbf{0.060}&\textbf{0.820}&\textbf{0.970}
				\\
				\Xhline{1.2pt}		
		\end{tabular}}
	\end{center}
	\caption{\textbf{Generalization to the SUN RGB-D dataset with models trained on the NYU-Depth-v2 dataset.} } 
\label{table5}
\end{table}
\begin{table}[htb!]
\begin{center}
	\renewcommand{\arraystretch}{1.3}
	\resizebox{1.0\columnwidth}{!}{\begin{tabular}{ c|| c c c c c c }
			\Xhline{1.2pt}
			Setting & PC & CIR & Abs Rel $\downarrow$ &  RMSE $\downarrow$  & ${\textbf{\rm{log}}_{\bm{{10}}}}$ $\downarrow$ & $\delta  < 1.25$ $\uparrow$\\
			\hline
			\hline				
			D&& & 0.095 & 0.334 & 0.041 & 0.922\\
			N$\mathcal{D}$&&& 0.092 & 0.324 & 0.039 & 0.926\\	
			N$\mathcal{D}$&\cmark&& 0.089 & 0.318 & 0.039 & 0.929\\	
			D$\&$N$\mathcal{D}$&\cmark&& 0.088 & 0.315 & 0.038 & 0.931\\
			D$\&$N$\mathcal{D}$&\cmark&\cmark& \textbf{0.087} & \textbf{0.311} & \textbf{0.038} & \textbf{0.936}\\	
			\hline
			D (planar)&& & 0.094 & 0.316 & 0.040 & 0.925\\
			N$\mathcal{D}$ (planar)&\cmark&& 0.088 & 0.301 & 0.038 & 0.934\\	
			D (non-planar)&& & 0.102 & 0.390 & 0.043 & 0.909\\
			N$\mathcal{D}$ (non-planar)&\cmark&& 0.104 & 0.394 & 0.044 & 0.908\\	
			\Xhline{1.2pt}			
	\end{tabular}}
\end{center}
\caption{\textbf{Results of ablation study}. D$\&$N$\mathcal{D}$: combined depth and normal-distance heads; D: depth head;  N$\mathcal{D}$: normal-distance head; PC: plane-aware consistency constraint; CIR: contrastive iterative refinement module. }
\label{table6}
\end{table}
In Figure~\ref{Fig7}, we visualize the refinement process of CIR
module. As the number of iterations increases, the locker
in the depth maps from the normal-distance head becomes
clearer, and the kitchen cabinet surface in the depth maps
from the depth head becomes more and more continuous.

We also compare the depth accuracy of normal-distance
head and depth head in planar and non-planar regions, respectively, and the results support our standpoint.

\section{Conclusion}
In this paper, we introduce a physics-driven deep learning framework for monocular depth estimation that leverages the planar information in
real-world 3D scenes. The framework contains two heads,
a normal-distance head and a depth head. The predicted
normal and distance are regularized by a developed plane-aware consistency
constraint. Besides, we develop a contrastive iterative refinement module to make full use of the strengths of these two heads according to the depth uncertainty. Extensive experiments indicate that the proposed method outperforms previous competitors on the NYU-Depth-v2, KITTI and SUN
RGB-D datasets. Existing works on monocular depth estimation ignore possible saturations in the input image. However, this is not always true in real-world applications~\cite{zheng2022neural}. It is desired to study 3D imaging and high dynamic range imaging together and this topic will be studied in our future research.

\textbf{Acknowledgments.} This work was supported in part by the National Natural Science Foundation of China under grant 51975029 and U1909215, in part by the A*STAR Singapore under MTC Programmatic Funds grant M23L7b0021, in part by the Key Research and Development Program of Zhejiang Province under Grant 2021C03050, in part by the Scientific Research Project of Agriculture and Social Development of Hangzhou under Grant No. 20212013B11, and in part by the National Natural Science Foundation of China under grant 61620106012 and 61573048.

{\small
\bibliographystyle{ieee_fullname}
\bibliography{egbib}
}

\end{document}